\documentclass[10pt,twocolumn,letterpaper]{article}

\usepackage{iccv}
\usepackage{times}
\usepackage{epsfig}
\usepackage{graphicx}
\usepackage{amsmath}
\usepackage{amssymb}
\usepackage{multirow}

\usepackage[pagebackref=true,breaklinks=true,letterpaper=true,colorlinks,bookmarks=false]{hyperref}

\iccvfinalcopy 

\def\iccvPaperID{1811} 

\ificcvfinal\fi
\begin{document}

\graphicspath{ {fig/}{feat_vis/}{quant_comp/}{failure/}{qual/}}
\newcommand{\secref}[1]{Sec.~\ref{#1}}
\newcommand{\figref}[1]{Fig.~\ref{#1}}
\renewcommand{\eqref}[1]{Eq.~\ref{#1}}
\newcommand{\tabref}[1]{Tab.~\ref{#1}}
\newcommand{\cmm}[1]{\textcolor[rgb]{0,0.6,0}{#1}}
\newcommand{\AddImg}[1]{\includegraphics[width=0.12\textwidth]{#1}}
\newcommand{\rebuttal}[1]{{\textcolor{black}{#1}}}

\title{Weakly Supervised Learning for Salient Object Detection}

\author{Huaizu Jiang\thanks{Main part of this work was done when Huaizu Jiang was at Xi'an Jiaotong University.},~~~\etal
}


\maketitle

\begin{abstract}
Recent advances in supervised salient object detection has resulted in significant
performance on benchmark datasets.
Training such models, however, requires expensive pixel-wise annotations of salient objects.
Moreover, many existing salient object detection models assume that at least
one salient object exists in the input image.
Such an assumption often leads to less appealing saliency maps on the
background images, which contain no salient object at all.
To avoid the requirement of expensive pixel-wise salient region
annotations, in this paper,
we study weakly supervised learning approaches for salient object detection.
Given a set of background images and salient object images,
we propose a solution toward jointly addressing the salient object existence
and detection tasks.
We adopt the latent SVM framework and formulate the two problems together in a single integrated objective function: saliency labels of superpixels are modeled as hidden variables and involved in a classification term conditioned to the salient object existence variable, which in turn depends on both global image and regional saliency features and saliency label assignment.
Experimental results on benchmark datasets validate the effectiveness of
our proposed approach.
\end{abstract}

\section{Introduction}
{\color{red}Attention!!! There was a mistake of one-class SVM for salient object detection in the previous vision since the kernel trik is used for the training phrase only.}

Salient object detection, deriving from classical human fixation
prediction~\cite{itti1998model},
\rebuttal{aims} to separate the entire salient object(s) that attract most of humans'
attention in the scene from the background~\cite{liu2011learning}.
Driven by applications of saliency detection in computer vision,
such as content-aware image resizing~\cite{avidan2007seam} and
photo collection visualization~\cite{wang2006picture},
many computational models have been proposed in the past decade.

\begin{figure}[t]
    \renewcommand{\arraystretch}{0.8}
    \renewcommand{\tabcolsep}{.4mm}
\begin{tabular}{cccc}
	\includegraphics[width=0.115\textwidth,keepaspectratio]{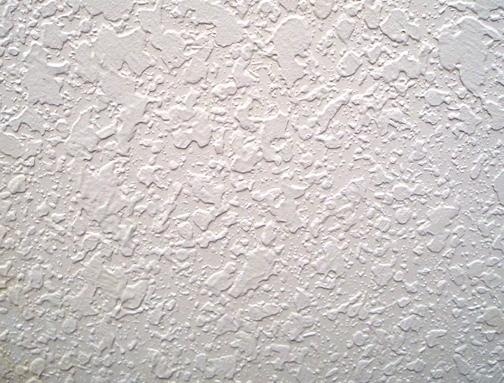}&
	\includegraphics[width=0.115\textwidth,keepaspectratio]{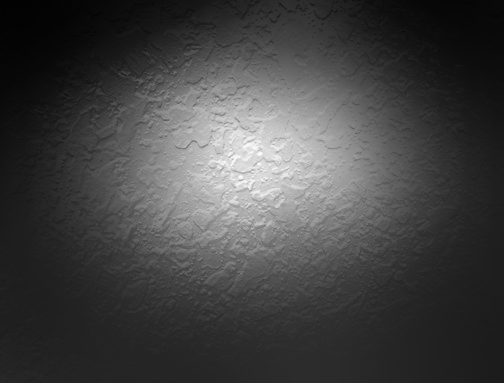}&
	\includegraphics[width=0.115\textwidth,keepaspectratio]{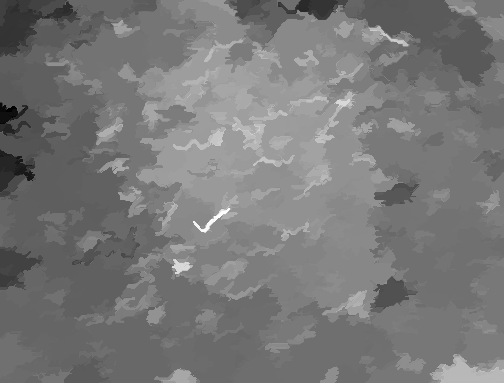}&
	\includegraphics[width=0.115\textwidth,keepaspectratio]{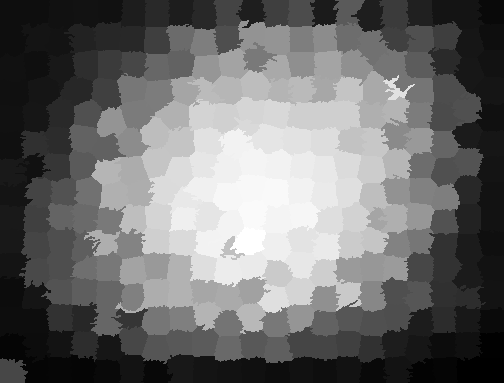}\\
	\includegraphics[width=0.115\textwidth,keepaspectratio]{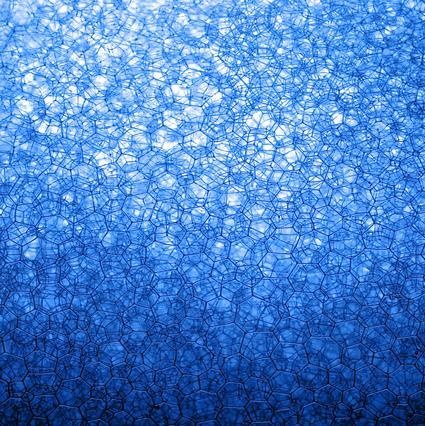}&
	\includegraphics[width=0.115\textwidth,keepaspectratio]{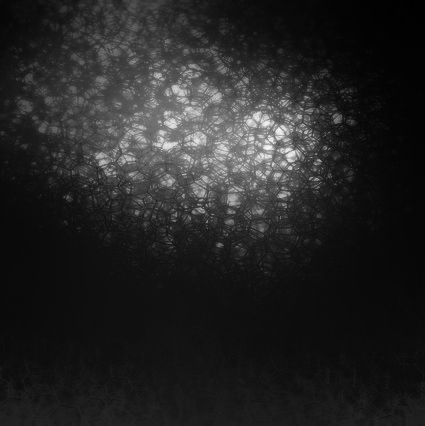}&
	\includegraphics[width=0.115\textwidth,keepaspectratio]{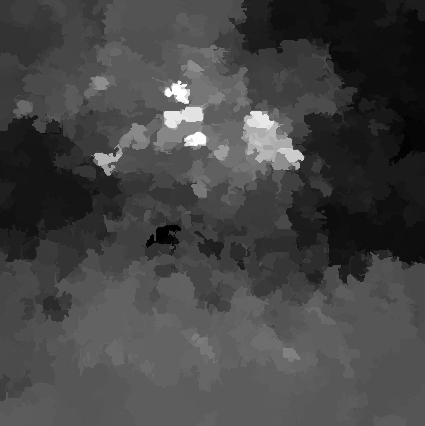}&
	\includegraphics[width=0.115\textwidth,keepaspectratio]{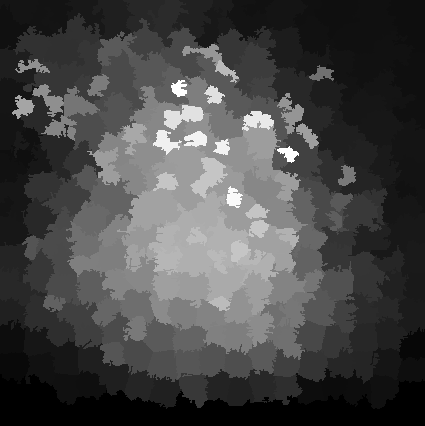}\\
	\includegraphics[width=0.115\textwidth,keepaspectratio]{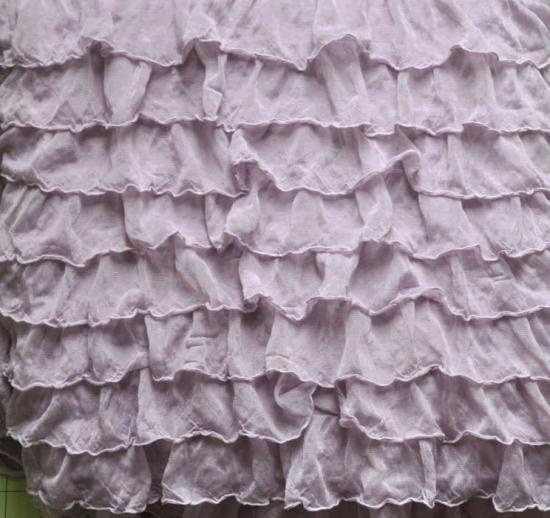}&
	\includegraphics[width=0.115\textwidth,keepaspectratio]{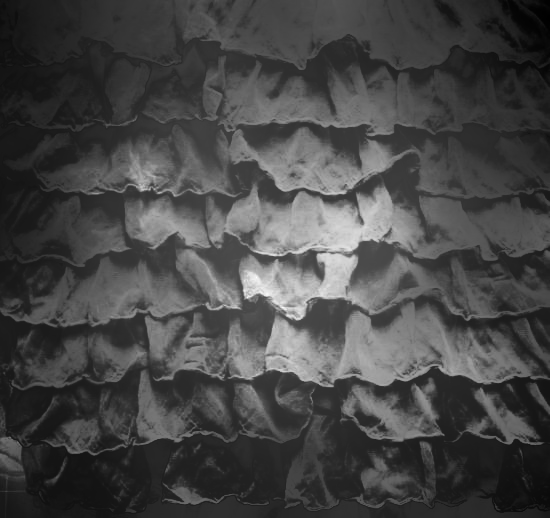}&
	\includegraphics[width=0.115\textwidth,keepaspectratio]{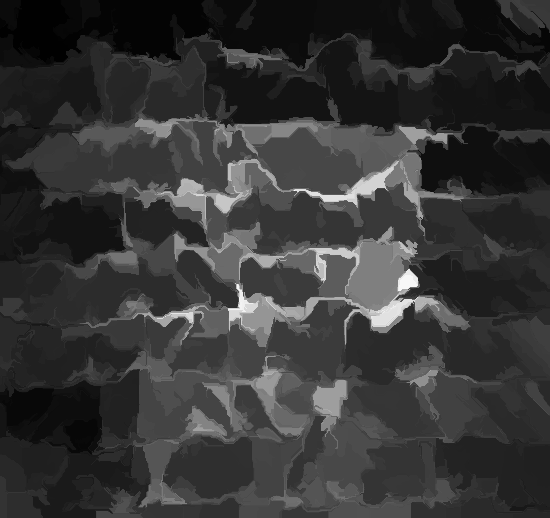}&
	\includegraphics[width=0.115\textwidth,keepaspectratio]{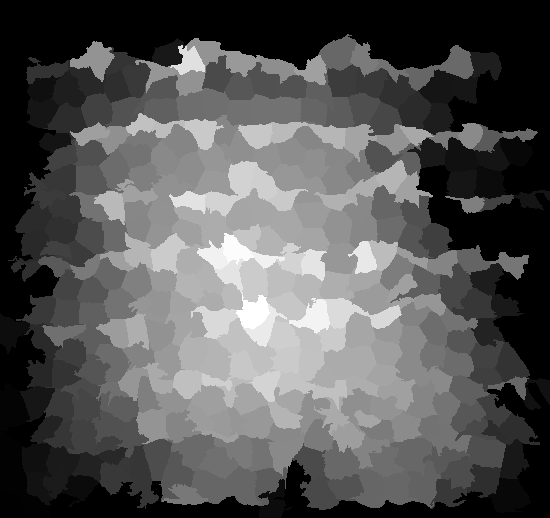}\\
	{\footnotesize(a) input} & {\footnotesize(b) SF~\cite{perazzi2012saliency}} & {\footnotesize (c) DRFI~\cite{JiangWYWZL13}} & {\footnotesize(d) RBD~\cite{zhu2014saliency}} \\
\end{tabular}
\caption{Saliency maps produced by three state-of-the-art models on the background images.}
\label{fig:intro_bg_imgs}
\end{figure}

There are two main motivations behind this paper.
On one hand, recent advances in supervised salient object detection has resulted in significant
performance on benchmark datasets~\cite{JiangWYWZL13}. 
Yet it is time consuming and tedious to annotate salient objects
in order to train a model.
On the other hand, it is usually assumed that at least one salient object exists
in the input image by most existing salient object detection algorithms (See~\cite{borjiTIP2014}).
However, as shown in~\figref{fig:intro_bg_imgs},
there exist \textit{background images}~\cite{wang2012salient},
where there are no salient objects at all. Based on this impractical assumption,
all of three state-of-the-art approaches
\cite{perazzi2012saliency,JiangWYWZL13,zhu2014saliency} produce inferior
saliency maps on background images.
To this end, we study how to utilize weakly labeled data to train salient object
detection models.
Given a set of background
images and salient object images, where we only have annotations of salient object existence labels, our goal is to train a salient object detection model.

In this paper, we propose a weakly supervised learning approach to jointly deal
with salient object existence and detection problems.
The input image is first segmented into a set of superpixels\footnote{
In this paper, we use the terms superpixel and region interchangeably.}.
Saliency labels of superpixels (\ie, foreground or background) are then modeled as
hidden variables in the latent structural SVM framework,
where the inference can be efficiently solved using the graph cut algorithm~\cite{Boykov04an}.
The training problem is built upon the large-margin learning framework to separate the
salient object images and the background images.
Our proposed weakly supervised approach is based on a set of unsupervised methods
\cite{ChengPAMI,perazzi2012saliency,YangZLRY13Manifold,zhu2014saliency}.
Compared with supervised approaches,
we do not require strong pixel-wise salient object annotations.
Furthermore, our approach is capable of recognizing the existence of salient objects.

Our main contributions therefore are two folds:
($i$) we propose a weakly supervised learning approach based on the latent structuralSVM framework,
instead of expensive salient object annotations;
($ii$) compared with conventional approaches,
our proposed approach is capable of jointly addressing salient object existence
and detection problems. Our approach performs better than most of unsupervised salient object detection models and is comparable with the best supervised approach.



\section{Related Work}\
\label{sec:related_work}
In this section, we briefly introduce related works in two areas:
salient object detection and weakly supervised learning for vision tasks.

\textbf{Salient object detection}. We refer readers to~\cite{borji2014salient,borji2012salient}
for a comprehensive review of salient object detection models.
Here, we 
briefly introduce some of the most related works. 

Visual saliency is usually related to the uniqueness, distinctiveness,
and disparity of the scene.
Consequently, most of existing works focus on designing models to capture the
uniqueness of the scene in an \textit{unsupervised} setting.
The uniqueness can be computed for each pixel in the frequency
domain~\cite{achanta2009frequency},
by comparing a patch to its most similar ones~\cite{goferman2012context},
or by comparing a patch to the average patch of the input image in the
principal components space~\cite{margolinmakes}.
Benefiting from image segmentation algorithms,
more and more approaches try to compute the regional uniqueness in a global
manner~\cite{ChengPAMI,perazzi2012saliency,borji2012exploiting},
based on multi-scale~\cite{jiang2011automatic} and hierarchical segmentations
of the image~\cite{yan2013hierarchical}.
Moreover, several priors about a salient object have been developed in recent years.
Since a salient object is more likely to be placed near the center of the image
to attract more attention (i.e., photographer bias),
it is natural to assume that the narrow border of the image belongs to the background.
Such a background prior is widely studied
\cite{wei2012geodesic,YangZLRY13Manifold,Jiang2013Saliency,li2013saliency}.
It is recently extended to the background connectivity prior assuming
that a salient object is less likely connected to the border area
\cite{zhu2014saliency,zhang2013boolean}.
In addition, generic objectness prior is also utilized for salient object detection
\cite{chang2011fusing,JiangLYP13UFO,jia2013category}.
Other priors include spatial distribution
\cite{perazzi2012saliency,ChengWLZVC13Efficient}
and focusness \cite{JiangLYP13UFO}.

There also exist \textit{supervised} salient object detection models.
The Conditional Random Field~\cite{liu2011learning,MaiNL13Aggregation}
and Large-Margin framework~\cite{lu2014learning} are adopted to
learn the fusion weights of saliency features.
Integration of saliency features can also be discovered based on the
training data using Random Forest~\cite{JiangWYWZL13},
Boosted Decision Trees (BDT)~\cite{mehrani2010saliency,kim2014salient},
and mixture of Support Vector Machines~\cite{khuwuthyakorn2010object}.

Our proposed weakly supervised approach is built upon the basis of the
feature engineering of several unsupervised approaches
\cite{ChengPAMI,perazzi2012saliency,YangZLRY13Manifold,zhu2014saliency}.
Compared with supervised approaches, however, our approach does not rely on strong
saliency annotations,
where we merely utilize the weak salient object existence labels of training images.
Moreover, our proposed latent salient object detection approach (\secref{sec:lssvm})
is capable of jointly addressing the salient object existence and detection problems.

\textbf{Salient object existence prediction}. In~\cite{wang2012salient},
the salient object existence prediction problem is studied as a standard binary
classification problem based on global saliency features of thumbnail images.
Zhang~\etal~\cite{zhang2015salient} investigate not only existence but also counting
the number of salient objects based on holistic cues.
In this paper, we focus on recognizing salient object existence.
By incorporating latent superpixels' saliency label in our approach,
better performance than~\cite{wang2012salient} can be achieved.
Moreover, salient object existence labels are used to train a weakly supervised
salient object detection model, predicting superpixels' saliency scores.

\textbf{Weakly supervised learning}. Visual data that are ubiquitously available
on the web are in nature weakly labeled, \eg,
images on Flickr and videos on YouTube with tags.
To leverage these data, weakly supervised learning methods are extensively studied
for vision tasks such as object detection~\cite{pandey11scene,deselaers12weakly},
concept learning~\cite{tang13discriminative},
scene classification~\cite{fergus07weakly,pandey11scene},
semantic image segmentation~\cite{vezhnevets12weakly}, etc.

In essence, our proposed approach is closely related to the work of visual concept
mining from weakly labeled data~\cite{Siva12in},
where we label the test data based on a strongly annotated negative training data.
Compared with~\cite{Siva12in},
our approach is more suitable for salient object detection.
In addition, our latent salient object detection based on the latent structural SVM is
closely related to the hidden~\cite{quattoni07hidden}
and max-margin~\cite{wang11hidden} conditional random fields.


\newcommand{\vw}{\mathbf{w}}
\newcommand{\bbR}{\mathbb{R}}
\newcommand{\calR}{\mathcal{R}}
\newcommand{\calX}{\mathcal{X}}
\newcommand{\calY}{\mathcal{Y}}
\newcommand{\calH}{\mathcal{H}}
\newcommand{\calI}{\mathcal{I}}
\newcommand{\calV}{\mathcal{V}}
\newcommand{\calE}{\mathcal{E}}
\newcommand{\vvs}{\mathbf{s}}
\newcommand{\vz}{\mathbf{z}}
\newcommand{\vI}{\mathbf{I}}
\newcommand{\vL}{\mathbf{L}}
\newcommand{\vD}{\mathbf{D}}
\newcommand{\vW}{\mathbf{W}}
\newcommand{\vV}{\mathbf{V}}
\newcommand{\vc}{\mathbf{c}}
\newcommand{\calG}{\mathcal{G}}

\section{Weakly Supervised Salient Object Detection}
\label{sec:lssvm}
In this section, we first present a weakly supervised approach for salient object detection based on the latent structural SVM framework (\secref{sub_sec:lssvm}). 
We then introduce saliency features used for salient object existence prediction and detection tasks (\secref{sub_sec:reg_sal_feat}).

\newcommand{\vh}{\mathbf{h}}

\subsection{A Latent Structural SVM Formulation}
\label{sub_sec:lssvm}
In this paper, we are interested in learning a model that can not only predict whether there exist salient objects in the input image but also where the salient objects (regions) are (if \rebuttal{they} exist). Our weakly annotated training data is composed of a set of images and their ground-truth annotations of salient object existence labels (\ie, salient object images vs. background images). Unlike supervised approaches~\cite{JiangWYWZL13,kim2014salient}, our approach does not need ground-truth annotations of regional saliency labels of the training samples. We call our approach \emph{weakly supervised} salient object detection since the supervision comes merely from the salient object existence annotations. It \rebuttal{is} worth exploring weakly supervised learning since it requires far less annotation effort than a supervised one.

Denote the input image as $I$, which consists of $N$ superpixels $\{r_i\}_{i=1}^N$.
Salient object existence label of the image is represented by a binary label $y\in\calY$,
where $\calY=\{0, 1\}$ denotes if there exist salient objects (0 for no existence).
Regional saliency labels of the image are denoted as $\vh=[h_i]_{i=1}^N$, where $h_i\in\calH=\{0, 1\}$ indicates the saliency label for the superpixel $r_i$ (0 is for background).

Given a set of training samples $\{(I_m, y_m)\}_{m=1}^M$, our goal is to learn a model that can be used to predict the salient object existence label $y$ as well as regional saliency labels $\vh$ of an unseen test image. To this end, we learn a discriminative function $f_{\vw}:\calI\times\calY\to\mathbb{R}$ over the image $I$ and its salient object existence label $y$, where $\vw$ are the parameters. During testing, we can use $f_{\vw}$ to predict the class label $y^*$ of the input image as $y^*=\arg\max_{y\in\calY}f_{\vw}(I, y)$. Due to lack of annotations, we model regional saliency labels $\vh$ as hidden variables in the latent structural SVM framework. We assume $f_{\vw}(I, y)$ takes the following form $f_{\vw}(I, y)=\max_{\vh}\langle\vw, \Psi(I, y, \vh)\rangle$, where $\Psi(I, y, \vh)$ is a feature vector depending on the input image $I$, its salient object existence label $y$, and regional saliency labels $\vh$.

We consider the global features $\Phi^e(I)$ of the input image $I$ to capture the salient object existence in a holistic manner as in~\cite{wang2012salient,zhang2015salient}. \rebuttal{Additionally, each superpixel $r_i$ is represented by two feature vectors 
$\Phi^f_i(I)$ and $\Phi^b_i(I)$,
modeling its negative
log-likelihood of belonging to the foreground and background, respectively.}
Their detailed definitions are introduced in~\secref{sub_sec:reg_sal_feat}.
To account for the spatial constraints of two adjacent superpixels that they tend to share the same saliency labels, we construct an undirected graph $\calG=(\calV, \calE)$. The vertex $j\in\calV$ corresponds to the saliency configuration of the superpixel $r_j$ and $(j, k)\in\calE$ indicates the spatial constraints of superxpixles $r_j$ and $r_k$. Finally, $\langle\vw, \Psi(I, y, \vh)\rangle$ is defined as follows,
\begin{align}
\langle\vw, &\Psi(I, y, \vh)\rangle = \sum_{a\in\calY}\delta(y=a)\langle \vw_a^e, \Phi^e(I)\rangle \notag\\
& +\sum_{a\in\{0, 1\}}\delta(y=a)\sum_{j\in\calV}\delta(h_j=1)\left(\langle \vw_a^s, \Phi^f_j(I)\rangle + w_a^f\right) \notag \\
& +\sum_{a\in\{0, 1\}}\delta(y=a)\sum_{j\in\calV}\delta(h_j=0)\left(\langle \vw_a^s, \Phi^b_j(I)\rangle + w_a^b\right) \notag \\
& - \sum_{(j, k)\in\calE}\delta(h_j\neq h_k) w^{p} \cdot v_{jk}.
\label{eqn:w_psi}
\end{align}
The model parameters $\vw$ are the concatenation of the parameters of all the factors in the above equation, \ie, $\vw=[\vw^e, \vw_a^s, , w_a^f, w_a^b, w^p]_{a\in\calY}$, where $w_a^f$ and $w_a^b$ are two prior terms for each region to be foreground and background,
respectively. 



 In the above formulation, both salient object existence prediction and detection problems are modeled together in a single integrated objective function. Salient object existence label does not only depend on the global image features $\Phi^e(I)$ in a standard classification term, but also on the regional saliency labels $\vh$ and features $\Phi_j^f(I)$ and \rebuttal{$\Phi_j^b(I)$}. Although we are at the same supervision level as existing supervised models of predicting salient object existence labels~\cite{wang2012salient,zhang2015salient}, regional saliency labels are taken into consideration as latent variables in our approach.

In turn, regional saliency labels $\vh$ are dependent on the salient object existence label $y$ as well. We learn two groups of model parameters for salient object detection on salient object images and background images, respectively. Moreover, we learn two prior terms $w_a^f$ and $w_a^b$ modeling the influence of salient object existence label $y$ on the latent salient object detection $\vh$. 
The last smoothness term encourages adjacent regions to take the same saliency label.
$v_{jk}$ captures the similarity of two neighboring regions $r_j$ and $r_k$.
It is defined as $v_{jk} = e^{-\frac{|\vc_j - \vc_k|^2}{2\sigma_c^2}}$,
where $\vc_j$ is the average color vector of the superpixel $r_j$ and parameter $\sigma_c$
is \rebuttal{set manually}.

\newcommand{\vp}{\mathbf{p}}
\newcommand{\vq}{\mathbf{q}}
\newcommand{\vu}{\mathbf{u}}

\renewcommand{\AddImg}[1]{\includegraphics[width=0.12\textwidth,keepaspectratio]{#1}}
\begin{figure*}[t]
	\centering
    \renewcommand{\arraystretch}{0.8}
    \renewcommand{\tabcolsep}{.4mm}
\begin{tabular}{ccccccc}
	\AddImg{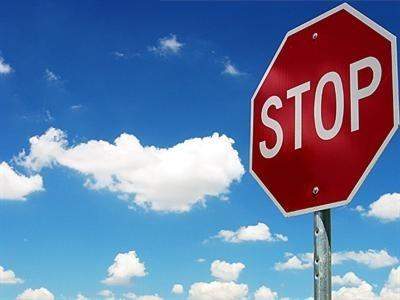}&
	\AddImg{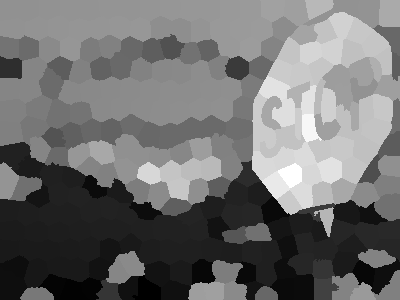}&
	\AddImg{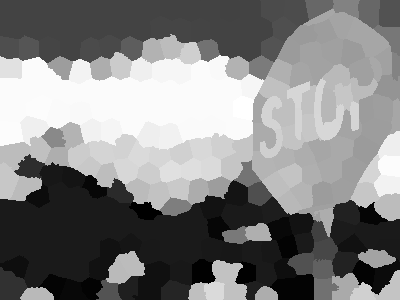}&
	\AddImg{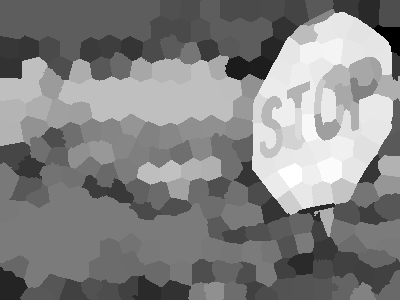}&
	\AddImg{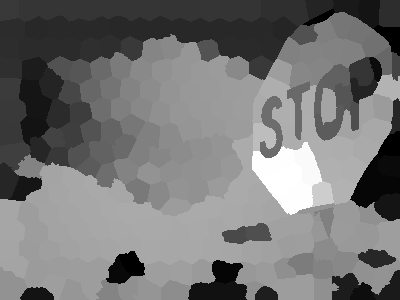}&
	\AddImg{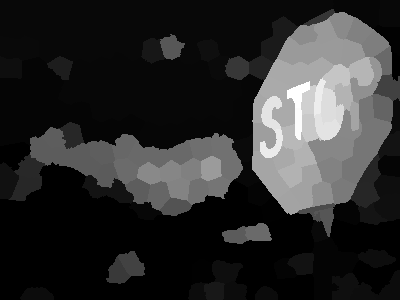}&
	\AddImg{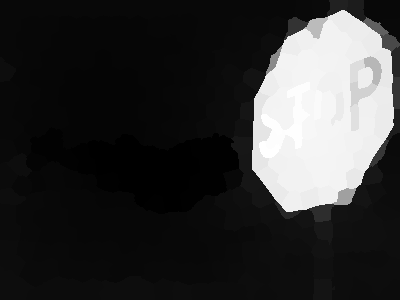}\\
	\AddImg{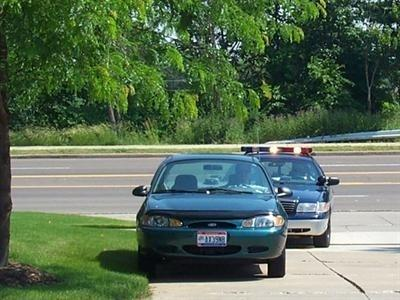}&
	\AddImg{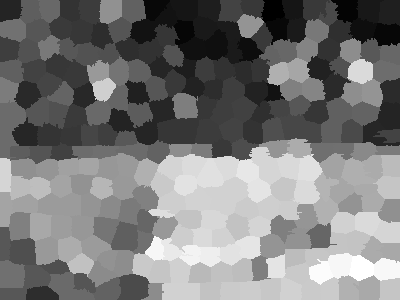}&
	\AddImg{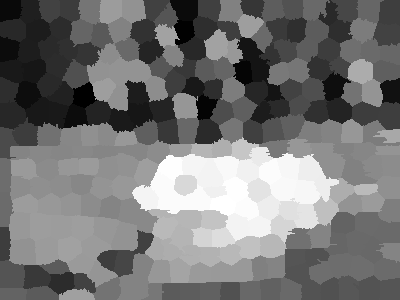}&
	\AddImg{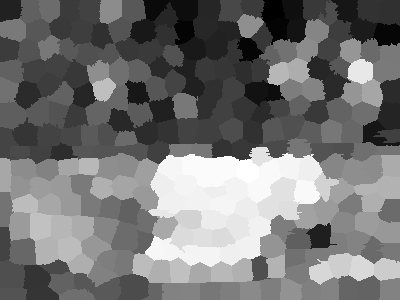}&
	\AddImg{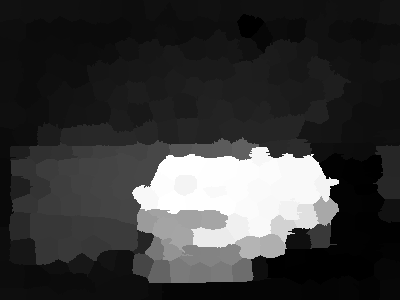}&
	\AddImg{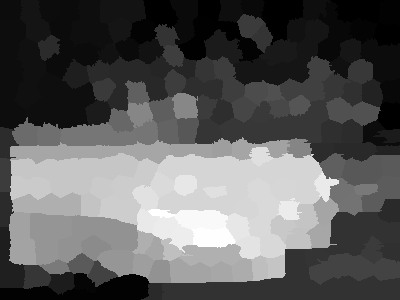}&
	\AddImg{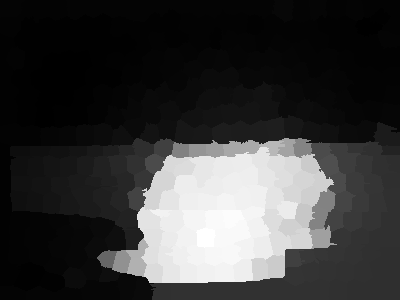}\\
	\AddImg{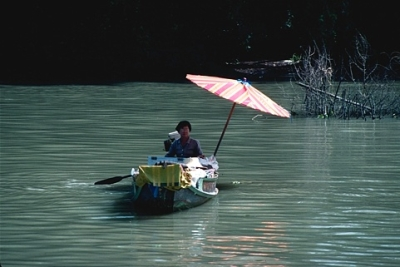}&
	\AddImg{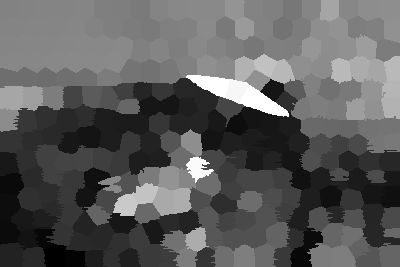}&
	\AddImg{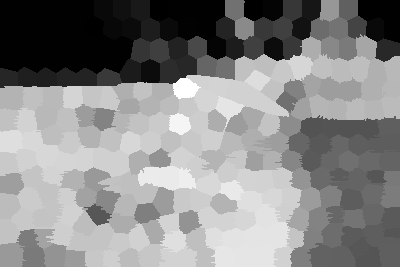}&
	\AddImg{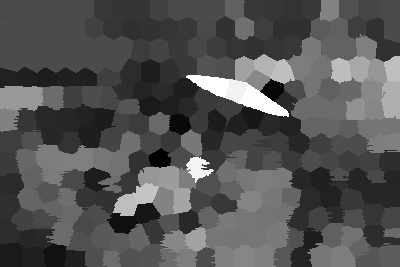}&
	\AddImg{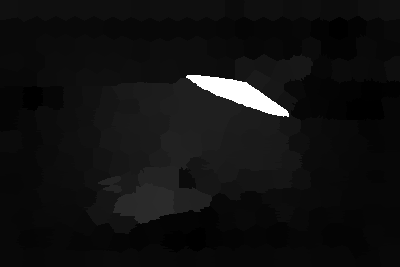}&
	\AddImg{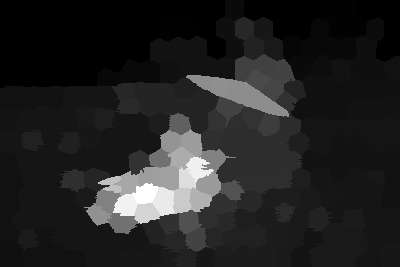}&
	\AddImg{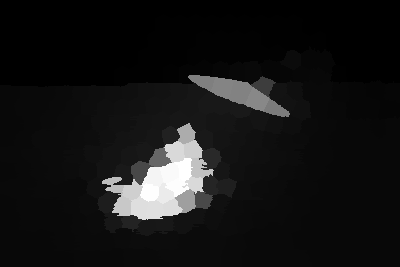}\\
	(a) & (b) & (c) & (d) & (e) & (f) & (g) \\
\end{tabular}
\caption{Illustration of saliency features computed on the Lab color histogram channel. From left to right: (a) input images, (b) global contrast, (c) spatial distribution, (d) backgroundness, (e) manifold ranking, (f) boundary connectivity, and (g) final saliency maps.}
\label{fig:feat_vis_nm}
\end{figure*}

\subsection{Saliency Features}
\label{sub_sec:reg_sal_feat}
In the past decade, reserchers have been mainly concentrating on designing various features
to describe salient objects.
Inspired by~\cite{ChengPAMI}, more and more research effort \rebuttal{is spent at the region level}.
In this paper, we consider the following five kinds of regional saliency features.

\textbf{Global contrast}. As studied in~\cite{ChengPAMI,perazzi2012saliency,borji2012exploiting},
the more distinct a region from others, the more salient it might be.
Regional global contrast $\Phi^{gc}_i(I)$ is computed by comparing the region $r_i$ to others, where nearby regions are given larger weights to determine the contrast value.

\textbf{Spatial distribution}. It is also an extensively studied saliency
feature~\cite{liu2011learning,perazzi2012saliency,ChengWLZVC13Efficient,borji2012exploiting},
indicating that the wider a region spreads over the image, the less salient it is.
Following~\cite{perazzi2012saliency}, we compute the spatial distribution $\Phi^{sp}_i(I)$
by computing spatial distances of the region $r_i$ with others, which are weighted by their appearance distances.

\textbf{Backgroundness}. Since the salient object is placed near the image center to
attract more attention,
the image borders $B$ are thus more likely belong \rebuttal{to} the background.
Following~\cite{JiangWYWZL13}, the regional backgroundness $\Phi^{bg}_i(I)$ is computed by examining
the region $r_i$ with respect to $B$ based on different appearance features.

\textbf{Manifold ranking}. In addition to directly comparing each region to the image border $B$,
a region's saliency score can also be
defined based on its relevance to $B$ via graph-based manifold ranking~\cite{YangZLRY13Manifold}.
Following~\cite{YangZLRY13Manifold}, we compute the ranking score for the region $r_i$ w.r.t each side of the image border $B$ and combine them together to get the final manifold ranking score $\Phi^{mr}_i(I)$.

\textbf{Boundary connectivity}. It is suggested in~\cite{zhu2014saliency}
that a salient region is less likely connected to the pseudo-background $B$.
To this end, the boundary connectivity score $\Phi^{bc}_i(I)$ of the region $r_i$ is defined as
the ratio between its spanning area and the length along the image border.


For robustness, we compute saliency features on different appearance channels including
average RGB, RGB histogram, average HSV, HSV histogram, average Lab, Lab histogram,
and Local Binary Pattern (LBP) histogram. Feature distances are computed as the $\chi^2$
distance for histograms and as absolute Euclidean distance for others.
Each dimension of the feature is normalized in the range $[0, 1]$.
Finally, we concatenate these five feature descriptors $\Phi^s_i(I)=[\Phi^{gc}_i(I),
\Phi^{sp}_i(I), \Phi^{bg}_i(I), \Phi^{mr}_i(I), \Phi^{bc}_i(I)]$.
In total, we obtain a 35-dimensional feature vector. See~\figref{fig:feat_vis_nm}
for examples of different saliency features. We refer readers to the original papers for more technical details.

Based on saliency features $\{\Phi^s_i(I)\}_{i=1}^N$, we adopt the same holistic manner as in~\cite{wang2012salient} to capture the existence of salient objects. We resize the pixel-wise saliency map resulting from each appearance channel of $\{\Phi^s_i(I)\}_{i=1}^N$ to $300\times300$ and
divide it into $5\times 5$ grids, concatenating the average saliency value in each grid
to form a global saliency feature vector $\Phi^{GS}(I)$. 
Additionally, we also consider the GIST descriptor~\cite{oliva2001gist} $\Phi^{GIST}(I)$, computed as a concatenation of averaged
responses of 32 Garbor-like filters over a $4 \times 4$ grid . Finally, we get a 1387-dimensional ($5\times 5\times 35 + 32\times4\times4$) feature vector $\Phi^e(I)=[\Phi^{GS}(I), \Phi^{GIST}(I)]$ to capture salient object existence.

We also define $\Phi^f_i(I)=-\log\left(1 - \Phi^s_i(I)\right)$
and $\Phi^b_i(I)=-\log\left(\Phi^s_i(I)\right)$, which can be regarded as the negative
log-likelihood of each region belonging to the foreground and background, respectively.
Since $\Phi^s_i(I)\in[0, 1]$, $\Phi^f_i(I)$ increases as it raises while $\Phi^b_i(I)$ decreases, indicating a region is more likely to be categorized as foreground with larger saliency feature values.

\section{Learning and Inference}
In this section, we introduce how to learn our model parameters $\vw$ from training samples (\secref{sub_sec:learning}) and how to infer both the salient object existence label $y$ and regional saliency labels $\vh$ given a test image (\secref{sub_sec:inference}).

\subsection{Large Margin Learning}
\label{sub_sec:learning}
Given a set of training samples $\{(I_m, y_m)\}_{m=1}^M$,
we find the optimal model parameters by minimizing the following regularized empirical risk~\cite{Do12regularized},
\begin{align}
\min_{\vw}~L(\vw) = \frac{\lambda}{2}||\vw||^2 + \frac{1}{M}\sum_{m=1}^M R_m(\vw),
\label{eqn:lssvm_objective_function}
\end{align}
where $\lambda$ controls the trade off between the regularization term
and the loss term.
$R_m(\vw)$ is a hinge loss function defined as
\begin{align}
R_m(\vw) = & \max_{y, \vh} \left(\langle\vw, \Psi(I_m,y,\vh)\rangle + \Delta(y_m, y, \vh)\right)\notag \\
 & - \max_{\vh}\langle\vw, \Psi(I_m,y_m,\vh)\rangle,
\end{align}
where the loss function $\Delta(y_m, y, \vh)$ is defined as follows
\begin{align}
\Delta(y_m, y, \vh) = \delta(y_m\neq y) + \alpha(y_m, \vh).
\end{align}
The first term is the 0/1 loss widely used for multi-class classification.
In addition, we introduce the second term to constrain the latent salient object
segmentation.
For a background image, its regional saliency labels should be all zeros.
For a salient object image, we resort to the pseudo-background prior~\cite{JiangWYWZL13} to treat all the saliency labels of regions in the border area of the image as zeros.
To this end, the second loss term can be written as
\begin{align}
\alpha(y_m, \vh) = \left\{\begin{array}{ll}
							\frac{1}{Z_0} \sum_{l=1}^N \beta_l \delta(h_l \neq 0 ), &\mbox{if $y_m=0$}, \\
							\frac{1}{Z_1} \sum_{l=1}^N \beta_l \delta(h_l \neq 0 )\delta(r_l\in B), &\mbox{if $y_m=1$}, \\
							\end{array}\right.\notag
\end{align}
where $\beta_l$ is the area of the region $r_l$.
$Z_0$ and $Z_1$ are normalization terms to ensure $\alpha(y_m, \vh)\in[0, 1]$.

\eqref{eqn:lssvm_objective_function} can be efficiently minimized using the bundle optimization method~\cite{Do12regularized}, which iteratively builds an increasingly accurate piecewise quadratic approximation of the objective function $L(\vw)$ based on its sub-gradient $\partial L(\vw)$. We first define
\begin{align}
\vh_y^* &= \arg\max_{\vh} \left(\langle\vw, \Psi(I_m,y,\vh)\rangle + \Delta(y_m, y, \vh)\right), \forall \rebuttal{m}, \forall y\in\calY, \notag\\
y_m^* &= \arg\max_{y\in\calY}\left(\rebuttal{\langle\vw, \Psi(I_m,y,\vh)\rangle} + \Delta(y_m, y, \vh_y^*)\right),
\end{align}
The sub-gradient $\partial L(\vw)$ can then be computed as
\begin{align}
\partial L(\vw) = \lambda\vw + \Psi(I_m, y_m^*, \vh_{y_m^*}^*) - \Psi(I_m, y_m, \vh_{y_m}^*).\notag
\label{eqn:loss_augment_optimization}
\end{align}
Given the sub-gradient $\partial L(\vw)$, the optimal model parameters can then be learned by minimizing~\eqref{eqn:lssvm_objective_function} using the method in~\cite{Do12regularized}.



\subsection{Inference}
\label{sub_sec:inference}
Given a test image $I$, we maximize~\eqref{eqn:w_psi}
to jointly predict its salient object existence label $y^*$ and regional saliency labels $\vh^*$ as follows,
\begin{align}
(y^*, \vh^*) = \arg\max_{y\in\calY, \vh} \langle\vw, \Psi(I, y, \vh)\rangle.
\end{align}
Since the search space $\calY$ of $y$ is small, we can iterate over all its possible values.
Given any $y\in\calY$, we utilize the max-flow algorithm~\cite{Boykov04an} to optimize
the~\eqref{eqn:w_psi} to get the optimal regional saliency labels. 

During training, we have to solve the loss-augmented energy
function~\eqref{eqn:loss_augment_optimization}.
Luckily, we can incorporate the loss of regional saliency labels
into the unary term of~\eqref{eqn:w_psi}.
Therefore, we can again utilize the max-flow algorithm~\cite{Boykov04an}
for efficient inference.

To output a saliency map, we diffuse the latent segmentation result of
salient object using the quadratic energy function~\cite{lu2014learning} as follows,
\begin{align}
\vz = \gamma(\vI + \gamma\vL)^{-1}\vI \vh,
\label{eqn:lssvm_diffusion}
\end{align}
where $\vz=[z_i]_{i=1}^N$. $z_i\in[0, 1]$ is the saliency value of the superpixel $r_i$. $\vI$ is the
identity matrix. $\vV=[v_{ij}]$ and $\vD=diag\{d_{11}, \cdots, d_{NN}\}$ is the
degree matrix, where $d_{ii}=\sum_{j}v_{ij}$. $\vL = \vD - \vV$ is the Laplacian matrix.

\renewcommand{\AddImg}[1]{\includegraphics[height=0.18\textwidth]{#1}}
\begin{figure*}[t]
    \renewcommand{\arraystretch}{0.8}
    \renewcommand{\tabcolsep}{.4mm}
    \begin{tabular}{cccc}
	   \AddImg{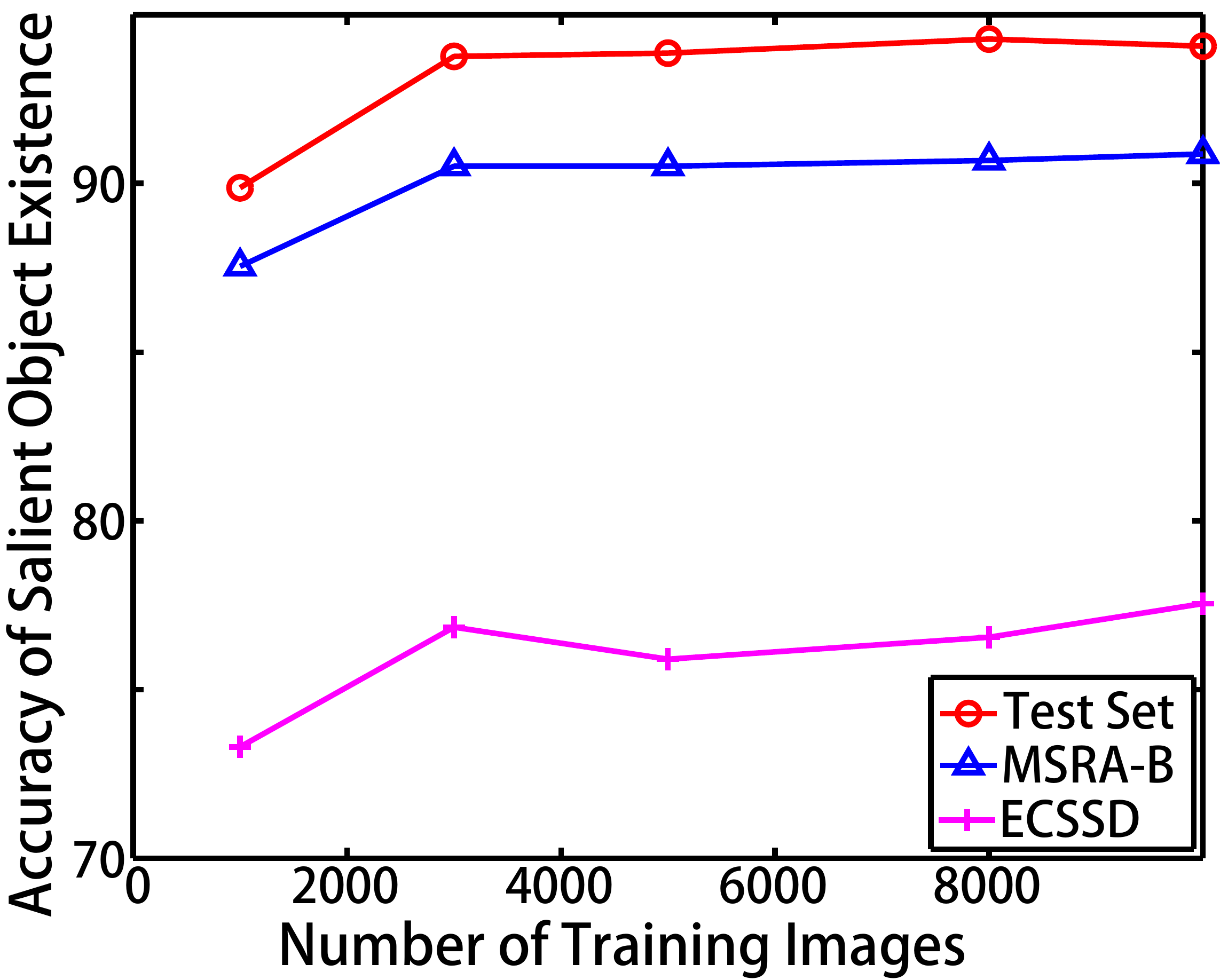}&
	   \AddImg{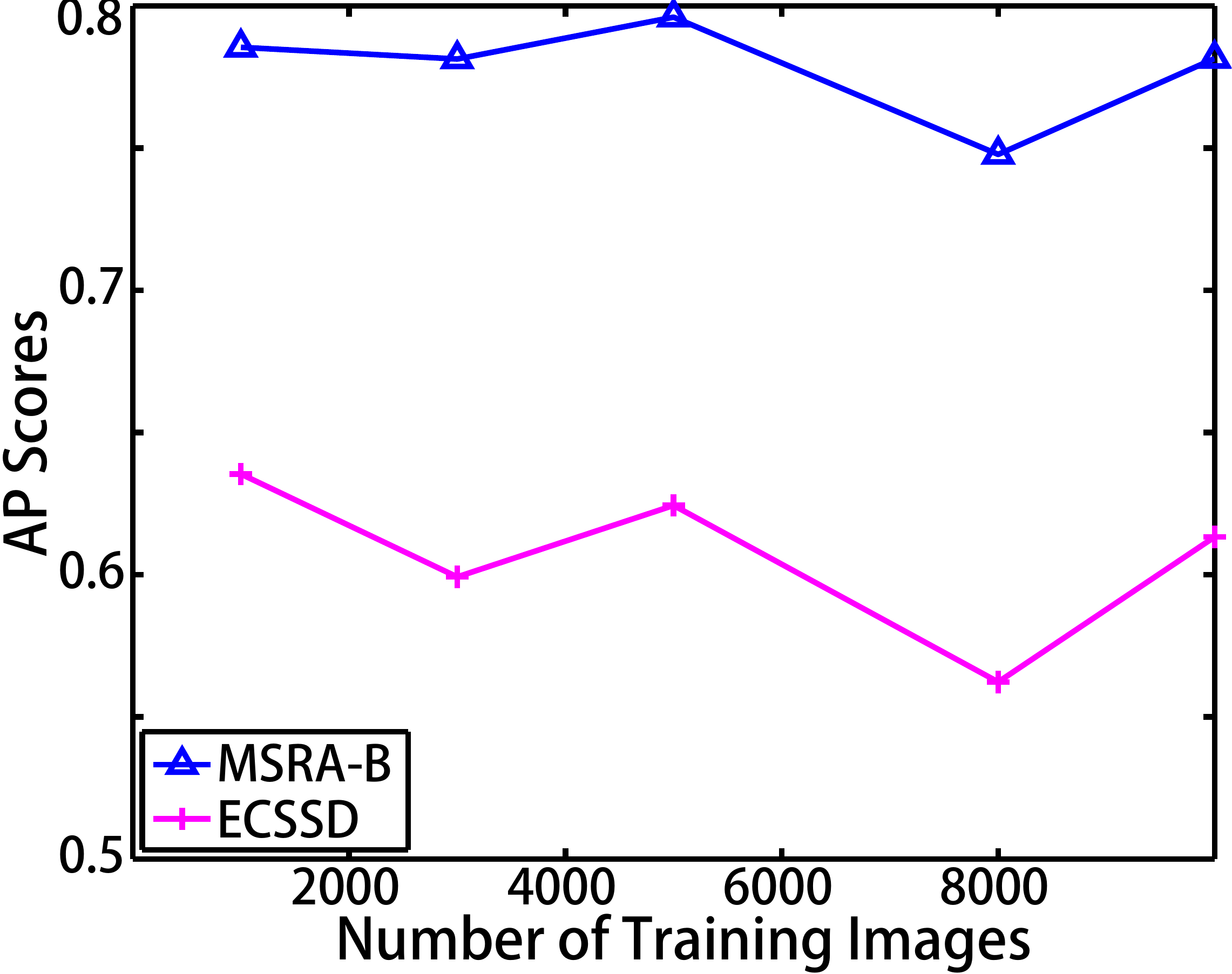} & \
        \AddImg{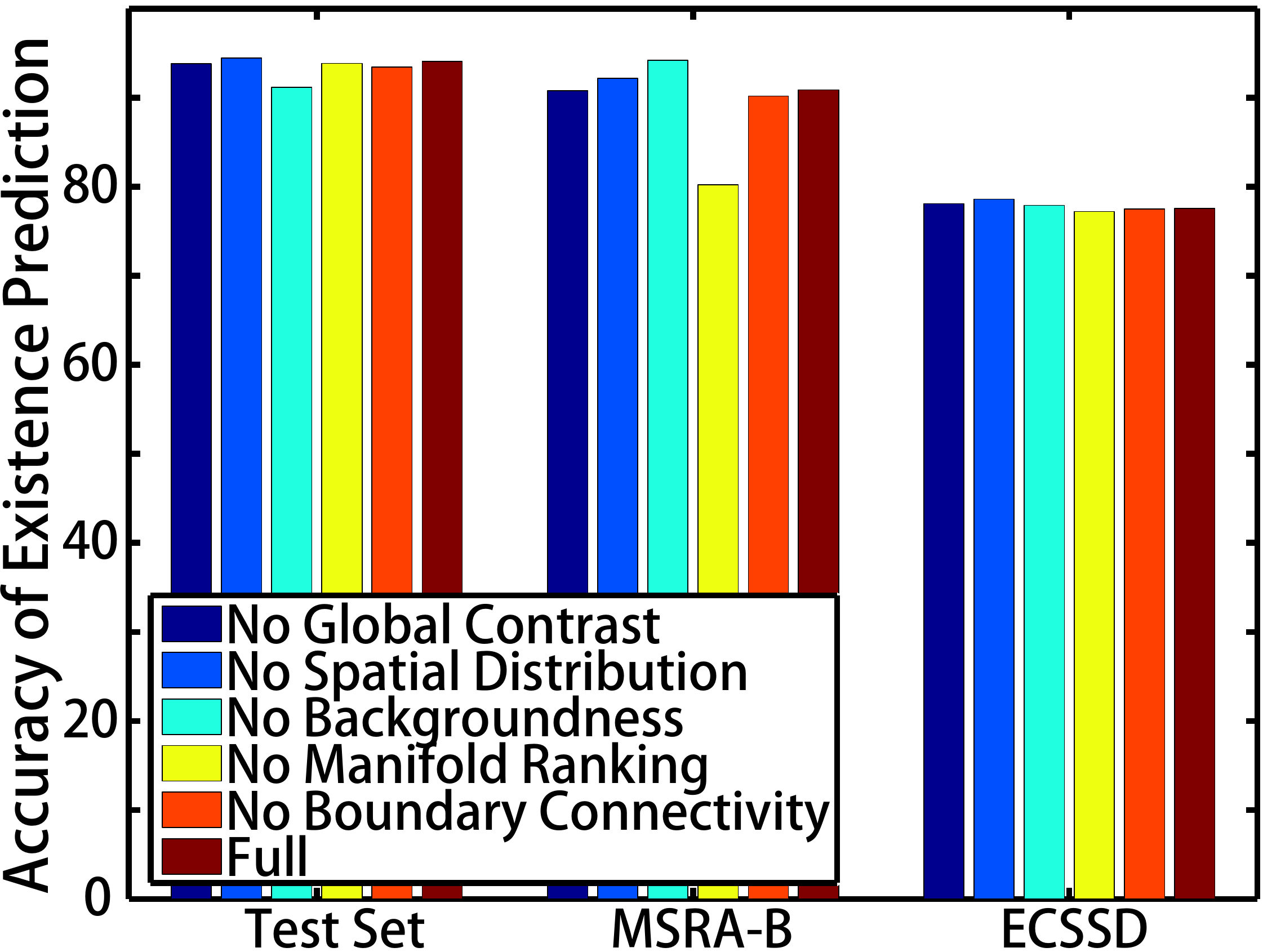}&
	   \AddImg{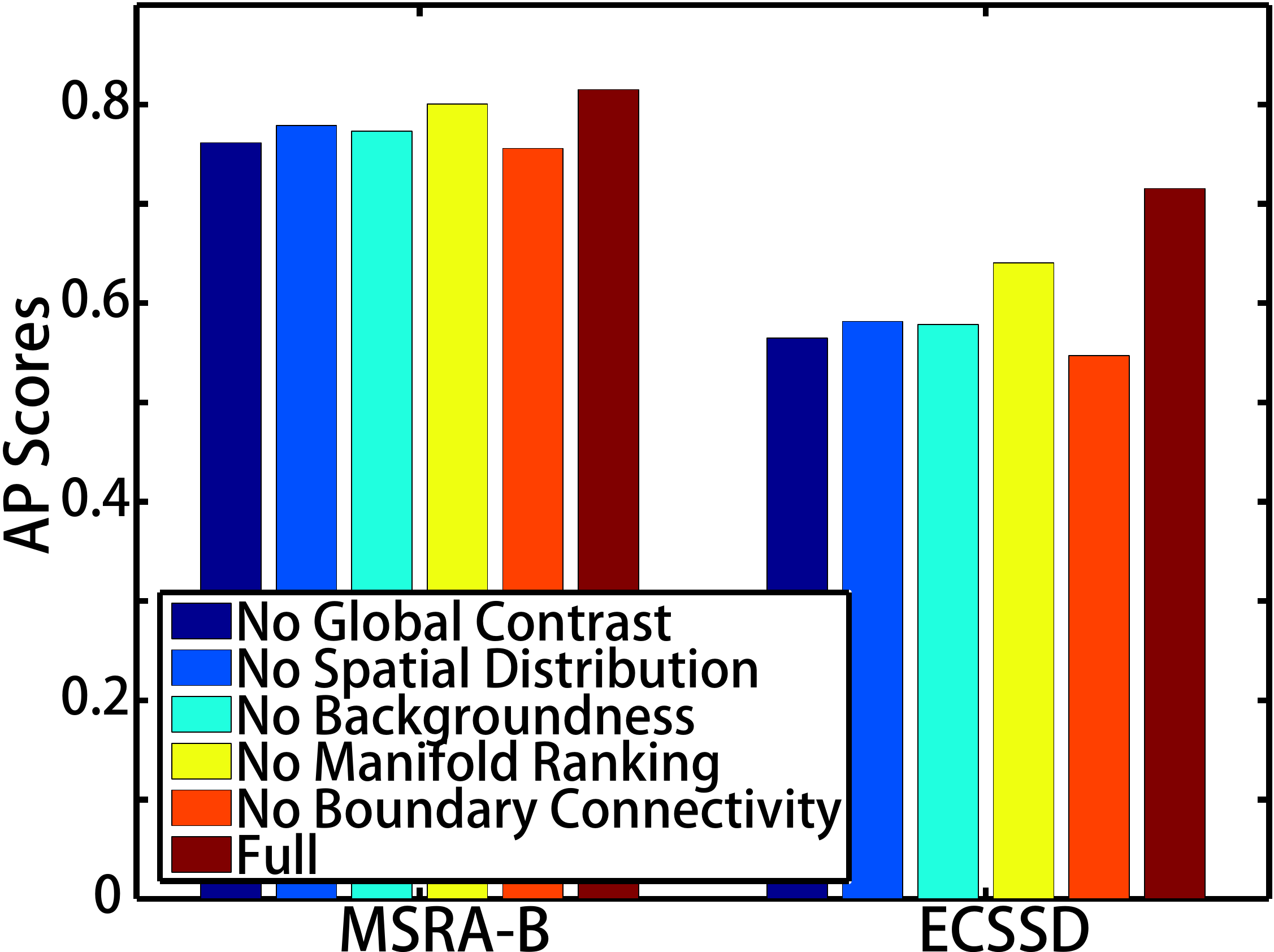}\\
	   (a) & (b) & (c) & (d)
    \end{tabular}
    \caption{Empirical analysis of our approach on the test set, MSRA-B,
        and ECSSD datasets.
        From top to bottom: (a)(b): accuracy of salient object existence prediction
        and AP scores of salient object detection versus different number of
        training images ($M$ in~\eqref{eqn:lssvm_objective_function}),
        (c)(d): accuracy of salient object existence prediction and AP scores of
        salient object detection versus different settings of feature combinations.
    }\label{fig:empirical_analysis}
\end{figure*}

\section{Experimental Results}
\label{sec:experiment}

\subsection{Setup}
Background images publicly available in the literature are only the thumbnail background image dataset~\cite{wang2012salient}. Images in this dataset, however, are of low resolution ($130\times130$). Since we are interested in images with common sizes (\eg, $400\times300$), this dataset is not suitable for our scenarios. To this end, we collect 6182 background images from the SUN dataset~\cite{xiao10sun}, describable texture dataset~\cite{cimpoi14describing}, Flickr, and Bing image search engines. We randomly sample 5000 background images to train our model and leave other 1182 images for testing. Additionally, we randomly sample 5000 images from the MSRA10K dataset~\cite{ChengPAMI} for training and 1237 images for testing. In total, we have 10000 images for training and 2419 for testing.

For the salient object detection task, we evaluate our proposed approach (LSSVM) on MSRA-B~\cite{JiangWYWZL13} and ECSSD~\cite{yan2013hierarchical} datasets with pixel-wise annotations. MSRA-B contains 5000 images with variations including natural scenes, animals, indoor scenes etc. There are 1000 semantically salient but structurally complex images in ECSSD, making it very challenging.

{\small
\begin{table}[t]
\caption{Taxonomy of different salient object detection algorithms based on supervision type and tasks that each method can solve. (Abbreviations unspvd. and spvd. denote unsupervised and supervised, respectively.)}
\label{tab:taxonomy_sal_obj_alg}
{\small
\begin{tabular}{c|c|c|c}
	\hline
	methods & supervision & task & pub. \& year \\
	\hline
	SVO~\cite{chang2011fusing} & unspvd. & detection & ICCV 2011\\
	CA~\cite{goferman2012context} & unspvd. & detection & CVPR 2010 \\
	CB~\cite{jiang2011automatic} & unspvd. & detection & BMVC 2011\\
	RC~\cite{ChengPAMI} & unspvd. & detection & PAMI 2015\\
	SF~\cite{perazzi2012saliency} & unspvd. & detection & CVPR 2012\\
	LRK~\cite{shen2012unified} & unspvd. & detection & CVPR 2012\\
	HS~\cite{yan2013hierarchical} & unspvd. & detection & CVPR 2013\\
	GMR~\cite{YangZLRY13Manifold} & unspvd. & detection & CVPR 2013\\
	PCA~\cite{margolinmakes} & unspvd. & detection & CVPR 2013\\
	MC~\cite{Jiang2013Saliency} & unspvd. & detection & ICCV 2013\\
	DSR~\cite{li2013saliency} & unspvd. & detection & ICCV 2013\\
	RBD~\cite{zhu2014saliency} & unspvd. & detection & CVPR 2014\\
	\hline
	DRFI~\cite{JiangWYWZL13} & spvd. & detection & CVPR 2013\\
	HDCT~\cite{kim2014salient} & spvd. & detection & CVPR 2014\\
	\hline
	\multirow{2}{*}{GS~\cite{wang2012salient}} & \multirow{2}{*}{spvd.} & existence & \multirow{2}{*}{CVPR 2012}\\
	\cline{3-3}
	& & localization & \\
	\hline
	\multirow{2}{*}{SOS~\cite{zhang2015salient}} & \multirow{2}{*}{spvd.} & existence & \multirow{2}{*}{CVPR 2015}\\
	\cline{3-3}
	& & counting & \\
	\hline
	\multirow{2}{*}{LSSVM} & weakly spvd. & detection & \multirow{2}{*}{}  \\
	\cline{2-3}
	 & spvd. + latent & existence & \\
	\hline
\end{tabular}
}
\end{table}
}

We compare our approaches with 14 state-of-the-art salient object detection models, including 12 unsupervised methods and 2 supervised models, which are summarized in~\tabref{tab:taxonomy_sal_obj_alg}. 
Following the benchmark~\cite{borji2012salient}, for quantitative comparisons, we binarize a saliency map with a fixed threshold ranging from 0 to 255. At each threshold, we compute Precision and Recall scores. We can then plot a Precision-Recall (PR) curve. To obtain a scalar metric, we report the average precision (AP) score defined as the area under the PR curve. Additionally, we also report the Mean Absolute Error (MAE) scores between saliency maps and the ground-truth binary masks.

\subsection{Empirical Analysis of Our Approach}

Here we empirically analyze our proposed approach on the test set, MSRA-B and ECSSD datasets. In particular, we quantitatively study the performance of both salient object detection and salient object existence prediction tasks by varying the following parameters.

\textbf{Number of Training Images}. As can be seen from~\figref{fig:empirical_analysis}(a), the latent structural SVM benefits from larger number of training samples, where the classification accuracy almost keeps increasing when more training images are adopted on all three datasets. However, according to~\figref{fig:empirical_analysis}(b), the performance of salient object detection does not always increase when more training samples are available. The reason might be that in contrast to the salient object existence, we have indirect (weak) supervision during training to constrain the salient object segmentation results.

\textbf{Feature Importance}. To measure the importance of features, we remove each kind of feature set and observe the performance variations on both tasks. In terms of salient object existence prediction, according to~\figref{fig:empirical_analysis}(c), the feature importance on three datasets are diverse. For instance, backgroundness is recognized as the most important on the test set while considered as the least critical one on MSRA-B.
Regarding the salient object detection tasks according to~\figref{fig:empirical_analysis}(d), the ranking of feature importance is consistent on MSRA-B and ECSSD. Features, from the most important to the least important are: boundary connectivity, global contrast, backgroundness, spatial distribution, and manifold ranking. It is worth noting that the full feature vector performs the best.


\renewcommand{\AddImg}[1]{\includegraphics[width=0.11\textwidth]{#1}}
\newcommand{\AddImgs}[1]{\AddImg{#1.jpg}&\AddImg{#1_sf.png}&\AddImg{#1_gmr.png}&
\AddImg{#1_dsr.png}&\AddImg{#1_rbd.png}&\AddImg{#1_hdct.png}&\AddImg{#1_drfi.png}&
\AddImg{#1_lssvm.png}\\}

\begin{figure*}[t]
    \renewcommand{\arraystretch}{0.8}
    \renewcommand{\tabcolsep}{.4mm}
    \centering
\begin{tabular}{ccccccccc}
    \AddImgs{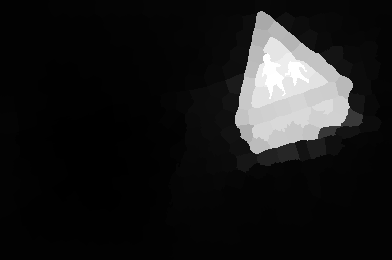}
	\AddImgs{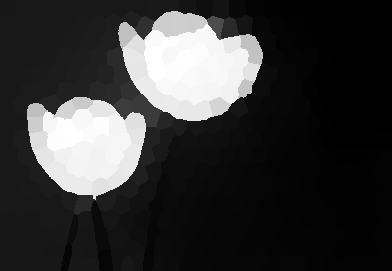}
	\AddImgs{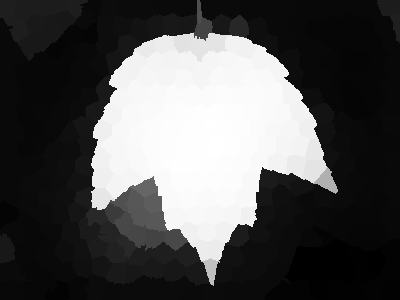}
	\AddImgs{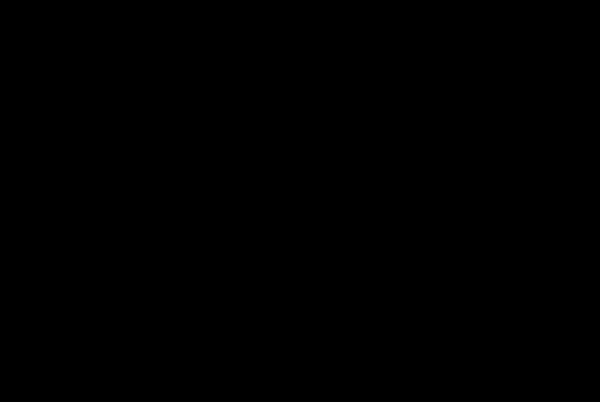}
	{\scriptsize (a) input} & {\scriptsize (b) SF~\cite{perazzi2012saliency}} &
    {\scriptsize (c) GMR~\cite{YangZLRY13Manifold}} & {\scriptsize(d)
    DSR~\cite{li2013saliency}} & {\scriptsize(e) RBD~\cite{zhu2014saliency}} &
    {\scriptsize(f) HDCT~\cite{kim2014salient}} & {\scriptsize(g)
    DRFI~\cite{JiangWYWZL13}} & {\scriptsize(h)LSSVM}\\
\end{tabular}
\caption{Qualitative comparisons of saliency maps produced by different approaches. From left to right: (a) input images, (b)-(g) saliency maps of state-of-the-art approaches, (h) saliency maps of our proposed approach LSSVM.}
\label{fig:qual_comp}
\end{figure*}

\subsection{Salient Object Existence Prediction}
\label{sub_sec:salObjExistencePrediction}
Here we quantitatively study our proposed approach in terms of the salient object existence prediction task. We compare our approach with three baselines, where we train a linear SVM, \rebuttal{two non-linear SVMs (using the $\chi^2$ and rbf kernels, respectively)}, and a Random Forest using our global image features $\Phi^e(I)$. As we can see \rebuttal{in~\tabref{tab:class_accuracy}}, by considering latent variables, our proposed approach (LSSVM) can achieve higher accuracy than the linear SVM. However, since both the rbf SVM, $\chi^2$ SVM and Random Forest are non-linear classifiers, they perform better than our approach. This motivates us that our approach may further benefit from non-linearly transforming our global features \rebuttal{(via a kernel function)}. \rebuttal{Therefore, we train a non-linear version of LSSVM (denoted as $\chi^2$ LSSVM), where we use the explicit feature mapping~\cite{vedaldi12efficient} to transform $\Phi^e(I)$ to approximate the $\chi^2$ kernel. As can be seen, benefiting from latent variables, its classification accuracy is still higher than its baseline ($\chi^2$ SVM) on both MSRA-B and ECSSD datasets. Moreover, it achieves the highest classification accuracy on the test set.}

Compared with the state-of-the-art approach in~\cite{wang2012salient}, our approach has two advantages, more powerful features and incorporation of latent saliency information. Though a non-linear classifier (Random Forest) is utilized in~\cite{wang2012salient}, as we can see from~\tabref{tab:class_accuracy}, our approach has higher classification accuracy on all datasets. Moreover, compared with~\cite{wang2012salient}, our approach is able to jointly address salient object existence and detection problems.

\begin{table}
\caption{Classification accuracy of different approaches on benchmark datasets. \rebuttal{(Updated.)}}
\centering
\renewcommand{\tabcolsep}{4.0mm}
\begin{tabular}{c|c|c|c}
	\hline
		 & Test Set & MSRA-B & ECSSD \\
	\hline
	linear SVM & 90.20 & 87.84 & 75.20 \\
	$\chi^2$ SVM & 93.14 & 90.80 & 81.80 \\
	rbf SVM &  95.37 & \textbf{93.16} & 82.90 \\
	RF &  92.24 & 91.52 & \textbf{84.50} \\
	\cite{wang2012salient} &  90.64 & 89.26 & 72.50 \\
	\hline
	LSSVM & 93.96 & 90.82 & 76.90 \\
	$\chi^2$ LSSVM & \textbf{95.58} & 92.54 & 79.90 \\
	\hline
\end{tabular}
\label{tab:class_accuracy}
\end{table}

\subsection{Salient Object Detection}
In this section, we compare our LSSVM approach with other state-of-the-art salient object detection approaches. 
Our LSSVM approach is designed to address the limit of conventional approaches, where they impractically assume that at least one salient object exists in the input image. 
\rebuttal{For more fairer comparisons, we introduce a two-stage scheme to make comparisons fairer. Specifically, we first predict the existence label of salient objects using the rbf SVM introduced in~\secref{sub_sec:salObjExistencePrediction}. If there are no salient objects, we output an all-black saliency map. Otherwise, we generate saliency maps using different approaches.}

In addition to MSRA-B and ECSSD benchmark datasets, we check performance of different approaches on the test set \rebuttal{consisting of 1237 salient object images and 1182 background images}. Since ground-truth annotations of background images are all-black images, only MAE scores are feasible to report on the test set. See~\tabref{tab:ap_mae} and~\figref{fig:pr_curve} for quantitative comparisons.


\rebuttal{Since an all-black saliency map is generated for the input that is classified as a background image, precision and recall scores are all zeros at all thresholds but 0 (the recall score is 1 when the threshold is 0, indicating all pixels are recognized as salient). This is why  
PR curves become flat when the recall approaches to 1.}

\rebuttal{We can see in~\figref{fig:pr_curve} that our approach PR curves are higher than others on most places. To this end, the linear version (LSSVM) outperforms other unsupervised and supervised approaches on both MSRA-B and ECSSD datasets in terms AP scores. Augmented with the explicit $\chi^2$ kernel feature mapping, better performance can be achieved, indicating that the salient object existence and detection problems can be mutually beneficial by modeling them in a unified framework. Specifically, $\chi^2$ LSSVM performs better than the second best method by 6.8\% (RBD) on MSRA-B and by 5.5\% (DRFI) on ECSSD. While the MAE scores are not as superior as the AP scores, $\chi^2$ LSSVM is ranked as the third best on both MSRA-B and ECSSD datasets. The reason why it performs inferior on the test set might be that our approach can not always produce all-black saliency maps for background images as other methods\rebuttal{\footnote{Recall that we produce an all-black saliency map if rbf SVM~recognizes an input as a background image.}}.}


In~\figref{fig:qual_comp}, we provide qualitative comparisons of our approach and other top performing approaches. As can be seen, our LSSVM approach can produce appealing saliency maps on images where salient objects touch the image border, although we utilize the background prior to extract regional saliency features and constrain the latent salient object detection. Moreover, on background images, our LSSVM approach generates near all-black saliency maps, clearly denoting no existence of salient objects.

\newcommand{\Coln}[2]{\multicolumn{#1}{|c}{{#2}}} 
\newcommand{\Colnn}[2]{\multicolumn{#1}{c|}{{#2}}} 
\newcommand{\Rows}[2]{\multirow{#1}{*}{#2}}
\newcommand{\first}{\color{red}\textbf}
\newcommand{\second}{\color{green}\textbf}
\newcommand{\third}{\color{blue}\textbf}
\begin{table}
\caption{AP and MAE scores compared with state-of-the-art approaches on different benchmark datasets, where supervised approaches are marked with bold fonts. The best three scores are highlighted with {\first red}, {\second green}, and {\third blue} fonts, respectively. \rebuttal{(Updated.)}}
    \tabcolsep3.0pt
\begin{tabular}{c|c|c|c|c|c}
	\hline
    \Rows{2}{} & \Colnn{2}{AP} & \Coln{3}{MAE} \\ \cline{2-6}
               & MSRA-B & ECSSD & MSRA-B & ECSSD & Test Set  \\
	\hline
\small{rbfSVM + SVO} & 0.631 & 0.458 & 0.333 & 0.388 & 0.212 \\
\small{rbfSVM + CA} & 0.512 & 0.390 & 0.241 & 0.326 & 0.101 \\
\small{rbfSVM + CB} & 0.652 & 0.483 & 0.184 & 0.275 & 0.111 \\
\small{rbfSVM + RC} & 0.672 & 0.506 & 0.135 & 0.233 & 0.093 \\
\small{rbfSVM + SF} & 0.607 & 0.473 & 0.168 & 0.270 & \first{0.052} \\
\small{rbfSVM + LRK} & 0.680 & 0.483 & 0.207 & 0.295 & 0.118 \\
\small{rbfSVM + HS} & 0.631 & 0.479 & 0.153 & 0.258 & 0.104 \\
\small{rbfSVM + GMR} & 0.709 & 0.517 & 0.126 & 0.235 & 0.085 \\
\small{rbfSVM + PCA} & 0.666 & 0.468 & 0.185 & 0.282 & \third{0.080} \\
\small{rbfSVM + MC} & 0.701 & 0.509 & 0.142 & 0.247 & 0.101 \\
\small{rbfSVM + DSR} & 0.694 & 0.524 & \second{0.119} & \second{0.229} & \second{0.076} \\
\small{rbfSVM + RBD} & \third{0.732} & 0.530 & \first{0.113} & \first{0.226} & 0.080 \\
\hline
\small{rbfSVM + \textbf{DRFI}} & \third{0.732} & \third{0.548} & 0.129 & \third{0.231} & 0.101 \\
\small{rbfSVM + \textbf{HDCT}} & 0.707 & 0.502 & 0.148 & 0.250 & 0.112 \\
\hline
\small{LSSVM} & \second{0.748} & \second{0.573} & 0.129 & 0.237 & 0.086 \\
\small{$\chi^2$ LSSVM} & \first{0.780} & \first{0.578} & \third{0.123} & \third{0.231} & 0.097 \\
	\hline
\end{tabular}
\label{tab:ap_mae}
\end{table}

\renewcommand{\AddImg}[1]{\includegraphics[width=0.4\textwidth]{#1}}
\begin{figure}[t]
    \renewcommand{\arraystretch}{0.8}
    \renewcommand{\tabcolsep}{.4mm}
    \centering
\begin{tabular}{c}
	\AddImg{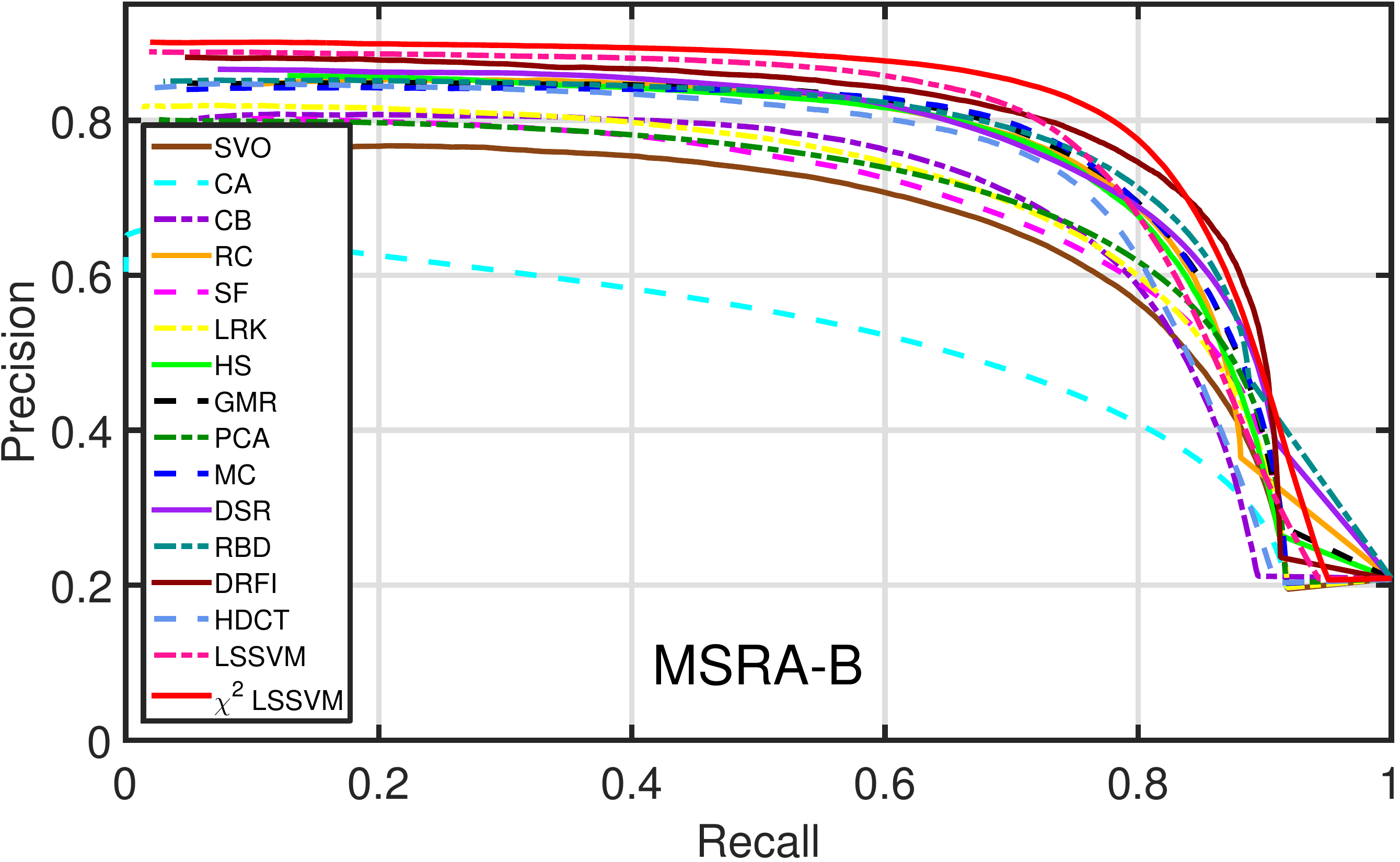}\\
	\AddImg{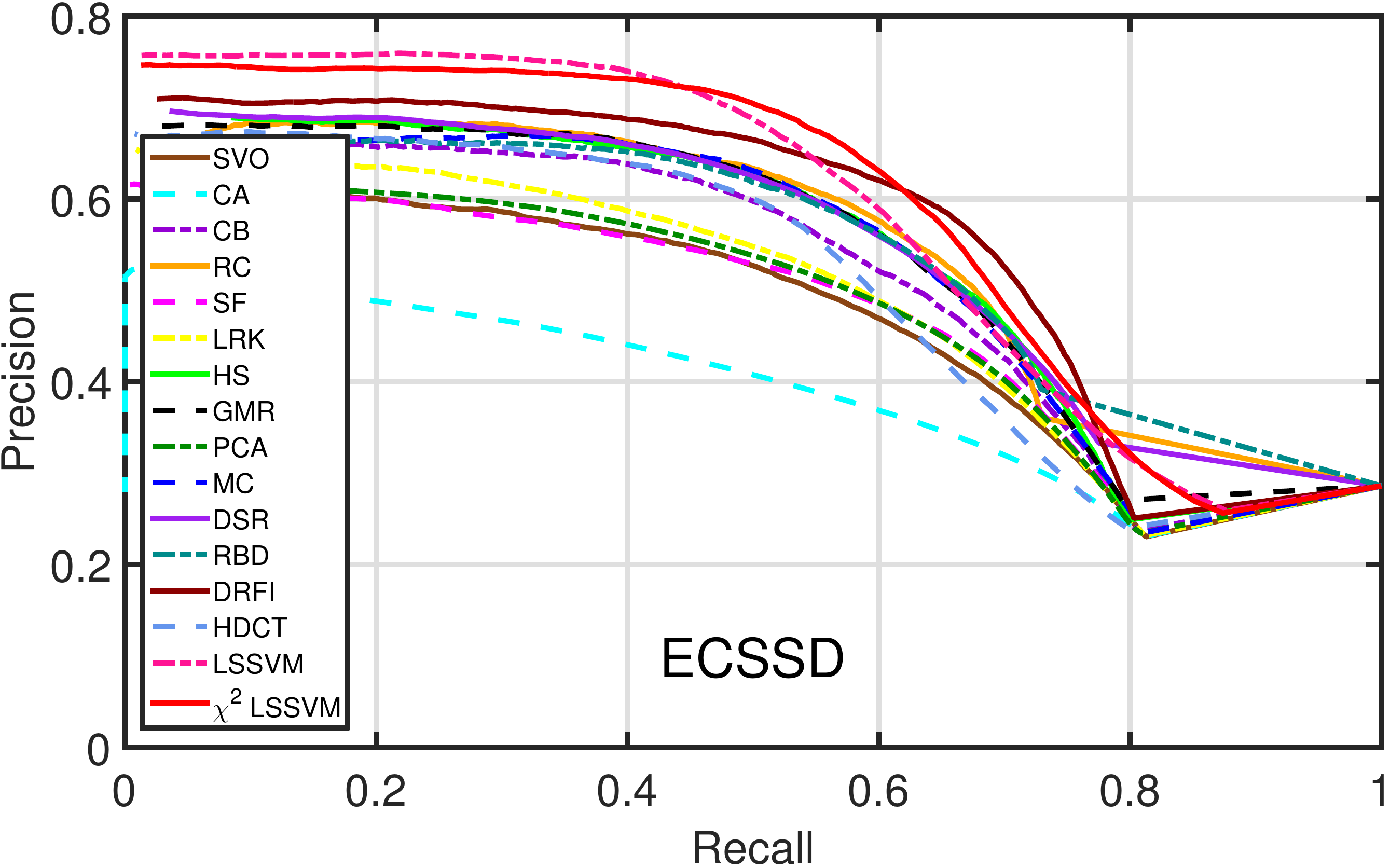}\\
\end{tabular}
\caption{Precision-Recall curves of different approaches on MSRA-B and ECSSD benchmark datasets. \rebuttal{(Updated.)}}
\label{fig:pr_curve}
\end{figure}

\rebuttal{On a PC equipped with an Intel i7 CPU (3.4GHz) and 32GB RAM, it takes about 12h to train our approach using MATLAB code and 0.5h to train the rbf SVM using C++. In testing, it takes around 3s to extract features. Our approach takes 0.02s for joint inference of the existence label and saliency map. In contrast, it takes 0.21s for the rbf SVM to predict the salient object existence (excluding feature extraction) and RBD takes 0.3s to output a saliency map.}

\subsection{Limitations}
Sometimes our approach makes incorrect classifications between salient object images and background images. See~\figref{fig:failure} for some failure cases. In the top row, the bird is hiding in the leaves, where the cluttered background and complex structure of the bird make the salient object detection difficult even for a human being at a first glance.
In the bottom row, textures of the image produce inferior saliency features, resulting in an incorrect classification.

\renewcommand{\AddImg}[1]{\includegraphics[width=0.11\textwidth]{#1}}
\renewcommand{\AddImgs}[1]{\AddImg{#1.jpg}&
	\AddImg{#1_bc.png}&
	\AddImg{#1_mr.png}&
	\AddImg{#1_lssvm.png}\\}
\begin{figure}[t]
    \renewcommand{\arraystretch}{0.8}
    \renewcommand{\tabcolsep}{.4mm}
    \centering
\begin{tabular}{cccc}
    \AddImgs{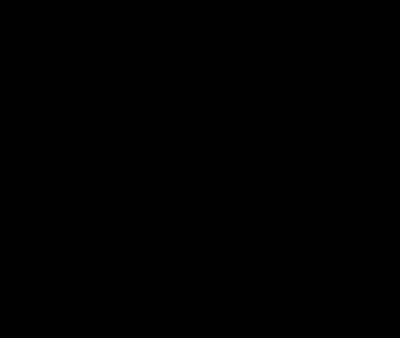}
	\AddImgs{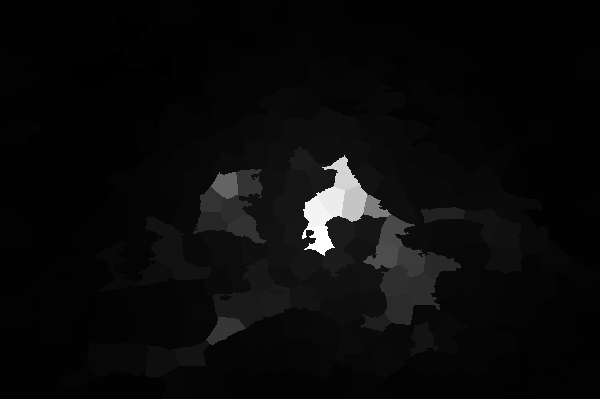}
	(a) & (b) & (c) & (d)\\
\end{tabular}
\caption{Failure cases of our LSSVM approach. Top row is a salient object image that is incorrectly recognized as a background image. Bottom row is a background image mis-classified as a salient object image. From left to right: (a) input images, (b)(c) saliency features of boundary connectivity and manifold ranking on the LAB histogram channel, and (d) saliency maps produced by our LSSVM approach.}
\label{fig:failure}
\end{figure}

\section{Discussion and Conclusion}
\label{sec:conclusion}
In this paper, we propose a weakly supervised learning approach for salient object detection based on the latent structural SVM framework using background images. Without any prior assumption of existence of salient objects, our approach is capable of jointly dealing with salient object existence prediction and detection tasks. Experimental results on benchmark datasets validate the effectiveness of our approach.

As a potential application, if we could recognize a background image, we no longer need to resort to complicated content-aware image resizing techniques (\eg~\cite{avidan2007seam}). Instead, standard bicubic interpolation method may be enough for background images shown in~\figref{fig:qual_comp}.

For future work, we plan to investigate more advanced global features, such as CNN features used in~\cite{zhang2015salient}, to further increase the accuracy of classification of salient object images and background images.

Since most existing approaches focus on unsupervised and supervised scenarios, we hope our work to draw attention of researchers on the weak supervision and make them realize the value of background images. We will release our code and background images for further research.


\small
{
\bibliographystyle{ieee}
\bibliography{egbib}
}

\end{document}